# Dual-Agent Framework for Cross-Model Verified Translation of Natural-Language Protocols into Robotic Laboratory Platform

Hyeonna Choi[a], Jung Yup Kim[b], Hyuneui Lim[a,*], and Seunggyu Jeon[a,*]

[a] Department of Bionic Machinery, Research Institute of AI Robot, Korea Institute of Machinery & Materials, 156, Gajeongbuk-ro, Yuseong-gu, Daejeon 34103, South Korea

[b] Department of Nano-devices & displays, Nano-convergence Manufacturing Research Division, Korea Institute of Machinery & Materials, 156, Gajeongbuk-ro, Yuseong-gu, Daejeon 34103, South Korea

* Corresponding authors.

**Hyuneui Lim**

Department of Bionic Machinery, Research Institute of AI Robot, Korea Institute of Machinery & Materials, 156, Gajeongbuk-ro, Yuseong-gu, Daejeon 34103, South Korea. Phone: +82-42-868-7106; E-mail: helim@kimm.re.kr; ORCID: 0000-0003-2692-647X

**Seunggyu Jeon**

Department of Bionic Machinery, Research Institute of AI Robot, Korea Institute of Machinery & Materials, 156, Gajeongbuk-ro, Yuseong-gu, Daejeon 34103, South Korea. Phone: +82-42-868-7902; E-mail: jsk1707@kimm.re.kr; ORCID: 0009-0002-3542-7859

**Abstract**

Biological experiment protocols are written in natural language, whereas automation systems rely on predefined control commands, creating a semantic gap that limits autonomous execution. Microplate-based automatic experiments are particularly challenging due to the need to simultaneously control well mapping, sample–reagent combinations, replicate placement, and parallel dispensing. This study proposes an agent-based protocol translation framework that converts natural-language microplate-based protocols into executable control commands for a robotic laboratory platform. A Parser Agent formalizes the natural-language protocol into a structured representation, and a rule-based mapping engine deterministically incorporates the operational constraints of the robotic laboratory platform to generate device-level control commands. A heterogeneous LLM Validation Agent verifies completeness, parameter accuracy, and execution order, and triggers a self-correction loop with structured feedback when errors are detected. A 7 Parser × 3 Validator sweep on randomly selected ELISA protocols evaluates how model scale and Validator type affect translation accuracy and pass rates under cross-model verification. The accuracy–latency trade-off is further verified by comparing the rule-based mapping of the proposed framework with LLM end-to-end direct mapping. Finally, Bradford assay–based protein quantification using a microplate was demonstrated on a robotic laboratory platform, validating end-to-end autonomous execution from natural-language protocols to real-world experiments. The proposed framework provides a flexible approach to narrowing the semantic gap between natural-language protocols and microplate based self-driving laboratories.

## 1. Introduction

Experimental automation in life science research has advanced continuously toward standardization, reproducibility, and the reduction of human error (Holland & Davies, 2020; King et al., 2009). Self-driving laboratories that integrate diverse instruments — including robotic arms, grippers, liquid handlers, and microplate readers — into a single unified environment have been widely established (Burger et al., 2020; Song et al., 2025; Steiner et al., 2019), and such systems contribute to the automation of repetitive experimental tasks, improvement of experimental consistency, and increased throughput. Among these, microplate-based experiments have emerged as a central experimental platform for high-throughput biomolecular analysis, immunoassays, and drug screening, enabling quantitative assays and automated large-scale experimentation (Groth & Cox, 2017; Klumpp-Thomas et al., 2021). However, conventional automation systems operate by relying on predefined workflows or control scripts (Ananthanarayanan & Thies, 2010; Bates et al., 2017). While real experimental procedures are commonly described in the form of natural-language protocols (Teytelman et al., 2016), the automation systems to directly interpret natural-language protocols is severely limited (Ivanov, 2024; Laurent et al., 2024). Consequently, researchers must manually translate natural-language protocols into workflows or commands compatible with their automation systems, and this translation process repeatedly becomes a bottleneck whenever new experiments are introduced or existing procedures are modified (Qu et al., 2026; Zhang et al., 2024). As a result, the semantic gap between natural-language protocols and automation systems remains one of the principal challenges in realizing autonomous experimental systems; in particular, research that automatically interprets natural-language protocols and translates them into executable commands for microplate-based experiments remains insufficient (Ivanov, 2024; Laurent et al., 2024; Ruan et al., 2024).

Microplate-based experimental protocols exhibit high variability in both structure and expression, and this variability introduces inherent technical difficulties in the process of converting natural-language protocols into executable structures for automation systems (Ivanov, 2024; Soldatova et al., 2014). Even within a single Enzyme-Linked Immunosorbent Assay (ELISA), numerous protocol variants exist depending on antibody combinations, the number of reaction steps, wash cycles, and detection methods, while reagent handling sequences, reaction conditions, and measurement approaches vary widely according to the assay objective and analyte of interest (Klumpp-Thomas et al., 2021; Ivanov, 2024; Laurent et al., 2024). This variability is equally evident not only in the structure of experimental procedures but also in the way they are expressed (Soldatova et al., 2014; Teytelman et al., 2016). Even for identical experimental procedures, the manner in which a protocol is written may differ depending on the researcher or experimental context; well layouts are designed in various configurations according to the placement strategies for calibrants, blanks, and samples (Klumpp-Thomas et al., 2021). Liquid-handling procedures vary with dispensing volumes, aspiration methods, and repetition counts, and tip-usage rules are defined differently according to cross-contamination prevention policies and multi-well dispensing strategies (Ananthanarayanan & Thies, 2010; Bates et al., 2017). The order in which experimental steps are performed is likewise not fixed. Steps such as shaking, heating, and incubation are included selectively, and the sequence and duration of each step vary across different protocols. Moreover, conditions such as temperature, time, and speed are described not only as quantitative values but also as qualitative expressions such as "room temperature." (Ivanov, 2024; Soldatova et al., 2014). This diversity in protocols and variability in expression introduce additional complexity in the automated translation process, and consequently, converting natural-language protocols for use in automation systems requires structural reasoning and contextual understanding rather than simple keyword matching (Kambhampati et al., 2024; Zhang et al., 2024).

Large language models (LLMs) have recently shown strong natural-language understanding across diverse tasks, making them promising tools for processing unstructured scientific text (Brown et al., 2020; OpenAI, 2023; Van Veen et al., 2024). Recent years have seen growing interest in applying LLMs to scientific document analysis and the automation of experimental procedures. Chemistry is a notable example, where LLMs have been used for experimental design and reaction planning (Boiko et al., 2023; Bran et al., 2024; Qu et al., 2026; Ruan et al., 2024). However, a single end-to-end LLM often struggles to produce reliable multi-step plans and cannot guarantee the executability of its outputs when translating natural-language protocols into executable commands (Kambhampati et al., 2024; Mirzadeh et al., 2025). Even recent approaches that incentivize reasoning capability through reinforcement learning at the training stage (DeepSeek-AI, 2025) still require a separate mechanism at inference time to guarantee executability. LLMs can produce hallucinations during generation, and when such hallucinations propagate directly into an automated experimental environment they result in execution errors (Farquhar et al., 2024; Ji et al., 2023; Xu et al., 2024). The hallucinations anticipated when using LLMs to translate unstructured biological experimental protocols can be distinguished into two forms. The first is execution hallucination, in which incorrect control commands are generated that fail to reflect equipment specifications or execution rules (Ahn et al., 2022; Liang et al., 2023; Yang et al., 2024). For example, commands that do not correspond to actual system behavior may be generated with respect to liquid-handling sequences, tip-usage rules, or instrument control conditions, and such errors lead directly to failures during experimental execution (Mon-Williams et al., 2025; Yang et al., 2024). To reliably translate natural-language protocols into device-level control commands, it is therefore necessary to introduce a deterministic intermediate translation step — such as a rule-based mapping engine — rather than using the LLM's free-form output directly as commands (Kambhampati et al., 2024; Liang et al., 2023). The second is reasoning hallucination, which refers to errors arising during the interpretation of experimental procedures, such as missing steps, ordering errors, and misinterpretation of conditions (Mirzadeh et al., 2025; Wang & Zhou, 2024). Such errors simultaneously undermine both the logical consistency of the experimental procedure and its experimental validity. In single-LLM approaches, the lack of an explicit mechanism to verify the executability of the generated protocol means that errors such as omitted steps, parameter inconsistencies, and failure to reflect equipment specifications may pass undetected and be forwarded directly to the execution stage (Farquhar et al., 2024; Huang et al., 2024). To reduce hallucinations and errors attributable to a single LLM, the agentic AI paradigm has been actively proposed (Chen et al., 2024; Qian et al., 2024; Wu et al., 2024). Agent-based systems structurally separate the processes of generation and verification through the collaboration of models with distinct roles, enabling error detection and correction via cross-verification between a generation agent and a validation agent (Chen et al., 2024; Park et al., 2025; Qian et al., 2024). This architecture functions as a key mechanism for improving the stability of reasoning and the overall reliability of the system (Chen et al., 2024; Song et al., 2025). These challenges therefore call for architectural strategies that move beyond single-LLM pipelines (Kambhampati et al., 2024; Yang et al., 2024).

This study proposes an agent-based autonomous protocol translation framework for converting unstructured experimental protocols written in natural language into executable commands for automated experimental systems. The contributions of this study are summarized as follows. First, this study presents an agent system that targets microplate-based biological experiments, which have not been sufficiently addressed in prior work, and that takes natural-language protocols as input and automatically generates executable control command sequences for the robotic laboratory platform. The proposed system can integrally generate the coordinated operations of multiple laboratory instruments, including plate transfer, liquid handling, and analytical instrument control, implementing the full pipeline from natural-language protocol to actual automated execution within a single framework. Second, to bridge the gap between natural-language interpretation and physical instrument control, a hybrid architecture that combines LLMs with rule-based control is proposed. The LLM is responsible for interpreting natural-language experimental protocols into structured representations, while the deterministic rule-based mapping engine generates executable control commands by reflecting physical constraints such as tip-usage sequences, instrument transfer paths, and coordinate transformations. This architecture structurally mitigates the hallucination issues that can arise in LLM-only approaches, and simultaneously ensures both the flexibility of semantic interpretation and the deterministic reliability of instrument control. Third, to secure the reliability of generated experimental procedures, a cross-verification structure based on heterogeneous models, separating the generation model from the validation model, is introduced. The proposed Validation Agent performs semantic verification of experimental procedures against criteria of completeness, parameter consistency, and step ordering; when an error is detected, the error information is fed back to the generation model so that the protocol is automatically regenerated. This structure mitigates the self-verification bias that can arise within the same model and improves the stability of the generation process. Finally, to validate the practical utility of the proposed framework, microplate-based liquid handling and Bradford assay-based total protein quantification experiments were conducted on the robotic laboratory platform, experimentally demonstrating the end-to-end autonomous executability from natural-language protocol to actual experimental execution. This architecture naturally gives rise to a human-in-the-loop collaborative paradigm that shifts the researcher's role from script writing to experimental design and supervision.

## 2. Proposed Framework

This study proposes a dual-agent protocol translation framework for converting natural-language experimental protocols into executable control commands for automated laboratory equipment. The proposed framework constitutes an integrated pipeline that combines semantic structuring of natural-language protocols, deterministic command mapping, and cross-model verification-based correction, and is designed to operate on the robotic laboratory platform (Fig. 1).

### 2-1. Robotic Laboratory Platform

The customized robotic laboratory platform (KIMM BioForge-1) on which the proposed framework operates is designed to perform microplate-based biological experiments automatically, carrying out plate transport, liquid handling, and analytical-instrument control within a single environment. It comprises three core modules (Fig. 2). The Plate Transporter, a SCARA robot, moves microplates between the Main Deck and the Analyzer Deck. The Liquid Handler, a gantry robot with a plate gripper, performs micro-level liquid handling and within-deck operations on the Main Deck — reagent aspiration and dispensing, tip attachment and removal, washing, plate transport between deck slots, and lid handling. The Plate Loader, a plate gripper, transfers plates among the Sample Plate, Lid Holder, and Microplate Reader slots in the Analyzer Deck and serves as the interface to the analytical instruments. The workspace is divided into two regions. The Main Deck has nine slots, each holding one fixture: Sample Plate, Magnet, Heating Plate, Tip Rack, Reagent Reservoir, Shaker, Tip Waste, Solution Waste, or Lid Holder. The Analyzer Deck has three slots housing the plate reader and associated instruments. The coordinate and numbering systems for each module and slot — as well as the Input/Output Stacker, sample-plate wells, reagent reservoir, and tip waste — are summarized in Supplementary Fig. 1 and serve as the basis for mapping well positions to instrument coordinates in the rule-based mapping stage. All operations are controlled through a standardized command structure, in which each command specifies the instrument module, the operation, and its parameters. These commands are the input to the upper-level control software that orchestrates the hardware, enabling consistent control across modules. The proposed framework targets this command structure as its translation output, converting natural-language protocols into executable command sequences.

### 2-2. Overall Dual-Agent Pipeline

To bridge the semantic gap between natural-language protocols and the automated instrument control system, the proposed framework constructs an integrated pipeline that combines three stages: protocol structuring, executable command generation, and cross-verification (Fig. 1). The system accepts a natural-language protocol document and microplate layout information as inputs. In the first stage, the LLM-based Parser Agent converts the natural-language protocol into a structured protocol representation. In the second stage, the rule-based mapping engine transforms this structured representation into a device-level control command sequence, operating as a deterministic execution layer that translates the LLM output into instrument-executable commands. In the third stage, the Validation Agent cross-verifies the generated command sequence against the original protocol; when errors are detected, a self-correction loop is triggered, iteratively refining both the structuring and command generation stages. The command sequence that passes final verification is executed on the robotic laboratory platform, completing an end-to-end workflow from natural-language protocol to actual automated experiment execution.

**2-3. LLM-Based Structuring of Natural-Language Experimental Protocols**

Natural-language biological experimental protocols contain unstructured narrative descriptions and diverse expression styles, making single-step conversion for automated systems difficult. To address this, the proposed framework designs a Parser Agent that converts natural-language protocols into a structured protocol representation usable within the automated execution pipeline (Fig. 3). The Parser Agent leverages an LLM to analyze the input protocol and transform it into a structured representation that captures experimental procedures, reagent handling steps, and environmental conditions. A constrained interpretation prompt is applied throughout this process to explicitly constrain the output format and experimental workflow structure. The output is designed to follow a predefined tag structure using tags such as <KIT_TITLE>, <MANUAL>, and <INSTRUMENT>, thereby normalizing the natural-language protocol into a consistent structured representation. The structured protocol is organized into two domains according to the agent performing each step in the experimental workflow. The Parser Agent explicitly separates each step of the natural-language protocol into user-performed and instrument-executed categories: the <MANUAL> domain covers steps carried out directly by the user—such as reagent dilution, plate coating and blocking, standard and sample dilution, and reservoir loading—while the <INSTRUMENT> domain contains steps executed by the automated instrument, such as liquid handling, incubation, washing, and absorbance measurement. This separation based on performing agent explicitly exposes the human–instrument task boundary that is otherwise intermixed in natural-language protocols, and serves as a safeguard ensuring that the subsequent rule-based mapping operates exclusively on instrument-executable steps. Furthermore, the constrained prompt is designed to restrict extraction to information present in the input document, thereby reducing the likelihood of generating erroneous information and preserving the ordering of experimental steps. This enables the generation of a consistent structured protocol representation across experimental protocols written in a variety of formats and expression styles. The structured protocol produced by the Parser Agent is subsequently used as input to the rule-based mapping engine, serving as an intermediate representation for converting natural-language experimental protocols into device-level control command sequences executable on the automated instrument.

**2-4. Rule-Based Compilation of Executable Control Commands**

Although the structured protocol produced by the Parser Agent captures the semantic structure of the experimental procedure, it is not in a form that can be executed directly on the automated instrument. To transform it into actual instrument control commands, the proposed framework designs a rule-based command mapping module that maps the structured protocol onto the execution command system of the robotic laboratory platform (Fig. 4). This module accepts the structured protocol as input and converts each experimental step into a device-level control command sequence; the conversion is performed based on a predefined set of rules. This layer operates as a deterministic execution layer that connects the structural information generated by the LLM to deterministic execution commands, and explicitly incorporates the physical constraints and operational rules of the automated instrument. The first component of this rule system is coordinate system mapping. The microplate is represented as a two-dimensional (x, y) grid, and users can define experimental roles—such as blank, calibrant, and sample—for each well. This user-defined plate layout is utilized during the rule-based mapping stage so that different reagent-handling and liquid-handling rules can be applied according to well type. The system also automatically groups wells to which the same reagent is applied, based on plate layout information, to optimize liquid-handling efficiency. Four consecutive wells in the same row are grouped into a 4-channel pipetting group so that multi-channel dispensing is applied preferentially; wells that cannot be handled this way are processed by single-channel operation. This strategy is designed to simultaneously improve processing speed and resource-use efficiency by reflecting the physical pipetting architecture of the automated instrument. The subsequent rules are defined by experimental operation type. Liquid handling rules convert operations such as reagent aspiration, dispensing, removal, and washing into the liquid-handling commands of the automated instrument. Tip management rules govern channel selection and tip attachment and replacement policies to prevent cross-contamination while minimizing tip consumption. Plate transport rules control the routing of plate movement between the Main Deck and the Analyzer Deck, and environmental control rules convert experimental environmental conditions—such as incubation, shaking, and temperature—into instrument control commands. At each experimental step, keywords and parameters are extracted and the corresponding rules are applied; the resulting execution parameters are then expanded into combinations of channel selection, pipette installation, aspiration, dispense, plate transport, and environmental control commands to form an executable command sequence (Fig. 5). Fig. 5 also illustrates the three-stage conversion process—Structured Protocol → Rule Assignment → Equipment Command Output—in which each step of the structured protocol is decomposed into a trigger keyword, an assigned rule number, and extracted parameters, before being expanded into a device-level control command sequence. This process reveals that a single step in the structured protocol is decomposed into anywhere from a few to several dozen commands at the instrument control level—for example, the Wash 4× step expands into 34 device-level control commands, and the Read OD step into 12—indicating that the rule-based engine serves as the key conversion layer that absorbs the granularity gap between semantic units and execution units. This separation design ensures the reliability and reproducibility of automated experiments by clearly distinguishing the uncertainty inherent in the natural-language semantic interpretation stage from the deterministic requirements of the instrument execution stage.

**2-5. Cross-Model Verification and Self-Correction**

During the conversion of natural-language experimental protocols into executable commands, errors such as parsing failures, parameter conversion errors, and execution order inconsistencies may arise. To detect and correct such errors, the proposed framework designs a Validation Agent-based cross-verification mechanism (Fig. 6). In the verification stage, the original protocol, the structured protocol, and the device-level control command sequence are all provided simultaneously as inputs. The Validation Agent is configured as a heterogeneous LLM-based verifier that is distinct from the model used for protocol structuring. This generator–verifier architecture is a core design choice aimed at blocking self-verification bias that may occur within a single model, and at mitigating reasoning errors and biases through cross-model verification. Verification is performed according to three criteria: completeness, parameter accuracy, and execution order. Completeness verification confirms that all experimental steps are reflected without omission, checking for missing reagent dispensing steps, washing cycles, reaction stages, and similar elements. Parameter accuracy verification checks that quantitative values—such as dispensing volumes, durations, and temperatures—have been correctly converted from natural-language expressions into instrument execution units; for example, it confirms that an expression such as "2 hours" has been correctly converted into the instrument control unit. Execution order verification evaluates whether the command sequence conforms to instrument specifications and control rules, verifying that liquid handling is performed after tip attachment, that plate movement follows a rational order, and that tip replacement policies are applied consistently. When errors are detected, the Validation Agent generates structured feedback containing the error type, its location, and correction guidance. This feedback is passed to the system to drive the self-correction loop: the structured protocol and command sequence are regenerated to incorporate the correction guidance, after which the same verification procedure is repeated iteratively. Iterations are performed within a pre-defined limit, and once all verification criteria are satisfied, the final commands are confirmed and executed on the robotic laboratory platform. The combination of cross-model verification and correction provides a mechanism that automatically detects and rectifies discrepancies between protocol interpretation and instrument execution, and constitutes a core component of the proposed framework for enhancing the reliability and operational stability of autonomous laboratory execution.

## 3. Experimental Method

The experimental design for quantitatively evaluating the proposed dual-agent protocol translation framework is as follows. The objects of evaluation are the operational outcomes of the framework and the contribution of its core component (the rule-based engine); definitions of the framework's components and operating mechanisms are not repeated here. The evaluation design is described in the following order: the protocol dataset, the ground truth construction procedure, the pool of models under evaluation, pipeline call conditions, quantitative metrics, experimental conditions, and the end-to-end demonstration on the robotic laboratory platform.

### 3.1 Test Protocol Dataset

The evaluation referenced 1,000 ELISA protocols collected from the web, from which 30 protocols differing in description style and format were selected for use in the evaluation. Collection targeted randomly selected commercial kit manual documents, with sources diversified to encompass both the expressive diversity and procedural variability of natural-language experimental protocols.

### 3.2 Ground Truth Construction

Two layers of ground truth were constructed for each protocol. The first layer is the structured protocol GT, which is the normalized structural representation that the Parser Agent is expected to produce; it conforms to the <KIT_TITLE>, <MANUAL>, and <INSTRUMENT> tag structure and a predefined step schema. The second layer is the device-level control command GT, which is the executable command sequence produced after the structured protocol passes through the rule-based engine and is decomposed into channel selection, pipette installation, aspiration, dispense, plate transport, and environmental control command units for execution on the robotic laboratory platform. By constructing the two layers of ground truth separately, the semantic structuring step of the Parser Agent and the deterministic mapping step of the rule-based engine can be evaluated independently. Labeling was carried out by multiple expert reviewers who each independently drafted a first-round ground truth, followed by a consensus review round for items on which disagreements arose. A minority of items that did not reach consensus were resolved through an additional review round, and the labeling results serve as the denominator and the matching criterion for Parameter Accuracy calculations in the evaluation phase.

### 3.3 Models

The candidate pool for the Parser Agent comprises seven models spanning a broad range of scale and hosting environment, from high-end cloud models to compact local models. The cloud-hosted models are GPT-5, GPT-4.1, o4-mini, GPT-4.1-mini, and GPT-4.1-nano; the open models, deployable locally or in the cloud, are llama-4-maverick and llama-3.3-70b. This range allows the effect of the self-correction loop to be examined across accuracy and latency regimes and across model sizes. Claude Sonnet 4.6 was adopted as the default Validation Agent. As an external model from a family distinct from the Parser pool (GPT and Llama series), it serves as the primary Validator and ensures generator–verifier heterogeneity. GPT-5 and llama-4-maverick were included as comparative Validators: GPT-5, the strongest cloud model within the Parser pool's family, tests whether model scale alone guarantees verification capability, and llama-4-maverick, an open-weight model deployable on-premise, tests the viability of a local Validator. Because the Parser and Validator pools partially overlap, some combinations are self-validation conditions; these were included alongside cross-model combinations for comparison on the same basis. The provider, hosting environment, version/snapshot, and release date of all models are summarized in Table 1 (specifications as of 2026-05-22). To isolate the effect of the model itself, all LLM calls used the same prompt template and the same call parameters (temperature, top-p, max tokens, and seed). For reproducibility, cloud calls recorded the endpoint, SDK version, and call timestamp with time zone, and local runs recorded the inference engine (vLLM or llama.cpp), GPU and memory specifications, and batch configuration. The timeout, retry count, and inter-call interval policy for rate-limit avoidance were fixed in advance and applied uniformly across all combinations.

**3.4 Pipeline and Self-correction loop**

The evaluation pipeline follows the flow of the proposed dual-agent framework exactly. A single attempt consists of Parser, rule-based mapping, and Validator in that order, and attempts are repeated for the same protocol until the Validator issues a pass judgment. In the present evaluation, the maximum number of attempts is capped at three, referred to as attempt 1, attempt 2, and attempt 3. Cases that do not pass after three attempts are classified as Final Fail. When the Validator detects an error in a given attempt, it generates structured feedback containing the error type, its location, and correction guidance, which is attached to the Parser input for the subsequent attempt. This attachment mechanism is intended to convey the Validation Agent's critical verification feedback to the subsequent attempt without loss. For the same (Parser, Validator, protocol) combination, the same seed and temperature are used to ensure reproducibility of results.

**3.5 Evaluation Metrics**

Parameter Accuracy is defined as a single metric that aggregates the correctly produced parameters across two layers — the structured protocol and the device-level control command sequence — while penalizing over-generation. The specific definition is as follows.

$$\text{Parameter Accuracy} = \frac{\max(C_{\text{struct}} - P_{\text{struct}}, 0) + \max(C_{\text{seq}} - P_{\text{seq}}, 0)}{T_{\text{struct}} + T_{\text{seq}}}$$

The two layers differ in their unit of content. In the structured layer, the meaningful content of a step is its parameters, such as volume, time, and temperature. In the command-sequence layer, by contrast, a single semantic step is expanded by the rule-based engine into a block of device-level commands spanning multiple rows — for example, a Wash 4× step expands into 34 rows of device-level commands and a Read OD step into 12 (Section 2-4) — so the executable content of a block resides in its rows, the individual commands actually run by the platform, as well as in its block-level parameters. Accordingly, the structured layer is counted in parameters, whereas the command-sequence layer is counted in rows and parameters.

The positive terms in the numerator, C_"struct" and C_"seq" , are the number of parameters that exactly match the GT within matched structured steps and the number of rows and parameters that exactly match the GT within matched command blocks, respectively, and represent the correct parameters produced by the model across the structured and command-sequence layers. The penalty terms, $P_{\text{struct}}$ and $P_{\text{seq}}$, correct for over-generation (false positives): $P_{\text{struct}}$ is the total number of parameters contained in over-generated structured steps, and $P_{\text{seq}}$ is the total number of rows and parameters in over-generated sequence blocks. The denominator terms, T_"struct" and T_"seq" , are the total number of structured parameters and the total number of rows and parameters in the command sequence according to the GT, forming a fixed denominator determined solely by the GT and independent of prediction length. The max(·, 0) operation is applied to each layer independently, clamping that layer's contribution to zero when its over-generation penalty exceeds its number of correct parameters; this prevents a heavily over-generated layer from producing a negative contribution while preserving the correct contribution of the other layer. In effect, Parameter Accuracy is the GT-normalized sum of each layer's correct parameters after subtracting its over-generation penalty (floored at zero per layer): because the denominator is fixed at the total GT parameter count, both omission (fewer correct parameters) and over-generation (a larger penalty) lower the score on the same GT-determined scale, regardless of prediction length. The matching unit is the step for structured output and the block for command sequences; matching is determined by the step or block type alone, and the key parameters are compared only after a match is established, so that completeness and parameter accuracy are assessed as independent axes. The Parameter Accuracy values on the y-axis of the graphs in the subsequent results analysis represent the overall accuracy as defined above. To compute Pass rate, the outcome of a protocol passing through the self-correction loop is classified into one of four categories. 1st Pass denotes cases passed by the Validator on attempt 1; 2nd Pass (Regenerated), cases in which attempt 1 fails and attempt 2 passes; 3rd Pass (Regenerated), cases in which attempts 1 and 2 fail and attempt 3 passes; and Final Fail, cases that do not pass after three attempts. The pass proportion of protocols is the proportion of each category among the 30 protocols per (Parser, Validator) combination, and decomposes how the self-correction loop contributes to recovery at each stage. Latency is the mean processing time per protocol for attempt 1, measured from the start of the Parser's LLM call to the completion of rule-based mapping; the Validator stage is reported separately. Because this measurement includes network latency from cloud API calls, the network conditions and call location at the time of measurement are recorded so that subsequent reproductions can be compared under the same conditions.

### 3.6 Test 1: Cross-Validator Sweep

To analyze the change in accuracy through repeated correction and the final pass rate after passing through the regeneration loop, all combinations of 7 Parser models and 3 Validator models — 21 combinations in total — are evaluated by passing 30 protocols through the self-correction loop for each combination. Per-attempt Parameter Accuracy and the final Pass category are recorded for each combination, and the same seed and temperature are used for the same combination and the same protocol to ensure reproducibility of results. This sweep enables quantitative analysis of how the effect of correction interacts with Parser model size and Validator selection.

### 3.7 Test 2: Accuracy–Latency Profiling

To quantify the relationship between Parser Agent model scale and accuracy, 30 protocols are translated by each of the 7 Parser models and the results are mapped onto an accuracy–latency plane. The vertical axis, Parameter Accuracy, represents the GT-referenced match rate for the Parser's semantic structuring and rule-based mapping results, while the horizontal axis, mean processing time per protocol, represents the average processing time per protocol from the start of the Parser's LLM call to the completion of rule-based mapping, reflecting both model scale and hosting environment. On this plane, two points are plotted together for each model — one for the condition without the self-correction loop (before) and one for the condition with it (after) — to visualize how correction shifts the accuracy–latency Pareto frontier.

### 3.8 Test 3: Proposed Framework vs. LLM End-to-End Mapping

To directly compare the hybrid architecture of the proposed framework against the LLM end-to-end approach, this experiment fixes the Parser to GPT-5. This choice is made because GPT-5 achieves the highest 1st Pass accuracy among the model pool, making it appropriate for isolating the effect of the mapping approach itself. Two comparison conditions are defined. In the Proposed Framework condition (Rule-Based), the Parser's output is passed through the rule-based mapping engine to be converted into device-level control commands; in the LLM End-to-End condition (LLM Direct Mapping), the same GPT-5 is prompted with instructions to output device-level control commands directly. Both conditions are evaluated on the same 30 protocols and the same ground truth data, and Parameter Accuracy and Latency are reported for each.

### 3.9 End-to-End Demonstration on the Robotic Laboratory Platform

For the end-to-end demonstration, the Parser was fixed to GPT-5 and the Validator to Claude Sonnet 4.6, the combination that produced the most stable cross-model verification in the sweep. Both demonstrations were executed physically on the platform, and only cases that passed at 1st Pass without regeneration are reported; call parameters were identical to those used in the sweep. Users interacted with the framework through a custom in-house chatbot, which submitted the natural-language protocol and returned the structured result with the Validator's feedback, and an open-source plate layout editor, modified to pass user-defined well assignments (blank, calibrant, sample) to the rule-based engine. Two procedures were demonstrated: a simple dispensing case and a Bradford assay measuring batch-to-batch variation in the total protein concentration of fetal bovine serum (FBS). The Bradford reagent and Bovine Serum Albumin (BSA) standard were obtained from Sigma-Aldrich (St. Louis, MO, USA), with a five-point BSA calibrant (5–25 μg/mL) and five batch-labeled FBS samples diluted to bring their protein concentration within the calibration range; the standard curve was fit by first-order regression against corrected optical density (OD) values.

## 4. Results & Discussion

### 4.1 Effectiveness of Iterative Correction across Validators

To analyze how the Validation Agent's iterative correction affects the translation accuracy of Parser models, parameter accuracy was measured across w/o regen. (before self-correction), attempt 1, attempt 2, and attempt 3 for seven Parser models — GPT-5, GPT-4.1, o4-mini, llama-4-maverick, GPT-4.1-mini, llama-3.3-70b, and GPT-4.1-nano — under each of three Validators: llama-4-maverick, GPT-5, and claude-sonnet-4-6 (Fig. 7). With the llama-4-maverick Validator, all Parser models reached attempt 3 with virtually no change from their baseline accuracy, and some smaller Parsers showed a slight dip at attempt 1 followed by a plateau (Fig. 7(a)); the self-correction loop produced no meaningful improvement for any Parser. The GPT-5 Validator showed a very similar pattern (Fig. 7(b)): changes across attempts were nearly absent, some Parsers declined modestly at attempt 1 before plateauing through attempt 3, and no gains were observed even for smaller Parsers such as GPT-4.1-nano. With the claude-sonnet-4-6 Validator, by contrast, parameter accuracy rose sharply after attempt 1 for all seven Parsers and then converged stably at attempts 2 and 3 (Fig. 7(c)). GPT-4.1-mini and llama-4-maverick climbed from approximately 0.4 at w/o regen. to 0.7–0.8 by attempt 3; llama-3.3-70b improved from 0.2 to 0.35–0.4; and GPT-4.1-nano rose from 0.069 to approximately 0.13. The two large Parsers were already near the ceiling: GPT-4.1 began at 0.931 and converged at approximately 0.95, while GPT-5 began at 0.998 and reached 1.0 across attempts, leaving little room for the correction loop to act.

The divergence in patterns across the three Validators shows that the effectiveness of correction depends not on the presence of extrinsic feedback but on the Validator's critical verification capability. The llama-4-maverick and GPT-5 Validators assigned passing judgments too liberally, whereas claude-sonnet-4-6 generated structured error feedback and thereby induced regeneration. The condition is therefore not heterogeneity per se but this verification capability, which makes Validator selection a key variable in dual-agent design. The self-validation combinations (GPT-5 × GPT-5; llama-4-maverick × llama-4-maverick) produced almost no effect, whereas the cross-model combinations, where a different Validator supplied the feedback, yielded a clear accuracy-recovery effect. Correction thus works when feedback comes from an external model, not from the model reviewing its own output, consistent with prior reports on the limits of self-correction (Huang et al., 2024). The mechanism of correction is tied to the lightweight nature of the framework. The present approach recovers the translation accuracy of small Parsers through inference-time validation and regeneration alone, without separate training or fine-tuning (cf. DeepSeek-AI, 2025). The recovery margin is limited for models with very low baseline performance (e.g., GPT-4.1-nano, 0.069), so the pathway applies to mid- and small-sized models with a certain level of baseline capability. This offers a practical route for deploying small models in experimental automation, where models operating alone have shown limitations on protocol-level tasks (Ivanov, 2024; Laurent et al., 2024).

### 4.2 Final Pass Rate under Cross-Verification

The analysis next examines how the cross-model verification structure operates across the stages of the regeneration loop. For each of the three Validators — llama-4-maverick, GPT-5, and claude-sonnet-4-6 — final outcomes were decomposed into four categories — 1st Pass, 2nd Pass (Regenerated), 3rd Pass (Regenerated), and Final Fail — and organized as the pass proportion of protocols for each Parser (Fig. 8). With the llama-4-maverick Validator, the outcome distribution split roughly in two between 1st Pass and Final Fail, with virtually no recovery in the regenerated categories (Fig. 8(a)); the shortfall of smaller Parsers accumulated almost entirely as Final Fail, so the contribution of the regeneration stage effectively vanished. The GPT-5 Validator showed the same pattern (Fig. 8(b)): large and mid-sized Parsers such as GPT-5, GPT-4.1, and o4-mini reached 1st Pass proportions of about 0.85 or above, whereas smaller Parsers such as GPT-4.1-mini, llama-3.3-70b, and GPT-4.1-nano remained at 0.05–0.6, with recovery in the regenerated categories negligible across all Parsers. With the claude-sonnet-4-6 Validator, by contrast, recovery through the regeneration loop was clearly observed (Fig. 8(c)). The 1st Pass proportion of large Parsers was already high, and for smaller Parsers a substantial share of protocols that failed at 1st Pass were recovered at the 2nd and 3rd Pass (Regenerated) stages. The cumulative pass rate of smaller Parsers such as GPT-4.1-mini, llama-4-maverick, and llama-3.3-70b thus rose markedly, and even GPT-4.1-nano, despite a very low 1st Pass proportion, reached a cumulative pass rate of approximately 0.4 after regeneration. Their Final Fail proportion was also markedly lower than under the llama-4-maverick and GPT-5 Validators.

Comparing the three panels, the contribution of the regeneration stage effectively disappears under the llama-4-maverick and GPT-5 Validators, while effective recovery occurs only with claude-sonnet-4-6. This matches the Validator-dependent pattern in Fig. 7 and shows that the framework's principle of pairing generator and verifier with different LLMs functions not through model heterogeneity alone, but through the Validator's critical verification capability. This finding bears directly on the hallucination problem. Rather than attempting to eliminate hallucination, which has been argued to be an intrinsic limitation of LLMs (Xu et al., 2024), the framework systematically filters and regenerates outputs that fail verification, providing a concrete recovery mechanism that complements detection-focused approaches (Farquhar et al., 2024). The confirmation bias that arises when one model both generates and verifies is well documented in the multi-agent literature (Chen et al., 2024; Wu et al., 2024); the consistently low Final Fail proportions under the claude-sonnet-4-6 Validator indicate that a cross-model structure mitigates this bias, provided the Validator has sufficient verification capability. The underlying mechanism is bias cancellation through diversity: a model with different training data and architecture does not share the error patterns that a single model would reproduce in both generation and verification, raising the likelihood that errors are caught. Compared with autonomous agents in related domains, which often rely on a single model or leave the verification structure implicit (Boiko et al., 2023; Singh et al., 2025; Song et al., 2025; Qu et al., 2026), the present framework makes cross-model verification explicit and adapts it to the constraints of physical experimental equipment.

### 4.3 Accuracy–Latency Trade-off across Parser Agents

In automated systems, model selection must account for processing latency as well as accuracy. A scatter plot of mean processing time per protocol against parameter accuracy (Fig. 9(a)) reveals the baseline trade-off: GPT-5, GPT-4.1, and o4-mini achieve high accuracy but require longer processing times, whereas GPT-4.1-nano and llama-3.3-70b are faster but start from lower accuracy. The correction results in Fig. 7 reshape this trade-off: because the effective accuracy of smaller Parsers rises markedly after correction while their low latency is unchanged, the achievable accuracy–latency frontier moves upward in the small-model region. Model selection is therefore governed by deployment requirements — a small Parser with correction in high-throughput settings, a large Parser where 1st Pass accuracy is critical, and a middle-ground option for mid-throughput cases balancing latency, accuracy, and cost. GPT-4.1 illustrates this middle ground: it reaches a 1st Pass accuracy of 0.93 at roughly one-quarter the processing latency of GPT-5 (Fig. 9(a)), placing it in the Pareto-dominant region and securing reliability at the 1st Pass stage while reducing reliance on the correction loop. These options are particularly relevant to self-driving laboratories, where the latency and data-privacy concerns of cloud-dependent LLMs are well known in Industrial IoT environments. Because a small Local LLM with correction can approach large-cloud accuracy while running on-premise, it keeps experimental data in-house and maintains operation under network failure — important where LLM inference must drive physical robot operations under latency constraints (Ahn et al., 2022; Liang et al., 2023).

### 4.4 Architecture Comparison: Proposed Framework vs. LLM End-to-End Mapping

Finally, the hybrid architecture is compared with an LLM end-to-end approach to assess the contribution of the rule-based engine. With the GPT-5 Parser held fixed, accuracy and latency were compared between the proposed framework, which maps structured protocols to device-level control commands through a deterministic rule-based engine, and an end-to-end approach in which an LLM performs the same mapping directly. The rule-based engine achieved both higher parameter accuracy and shorter processing time than direct LLM mapping (Fig. 9(b)), supporting a hybrid design that separates the LLM's semantic interpretation from the deterministic engine's enforcement of physical constraints. The accuracy loss in direct mapping stemmed primarily from execution hallucination — tip-usage or coordinate rules that did not reflect equipment specifications, and inaccurate step ordering — consistent with the known limitations of LLM-only approaches (Kambhampati et al., 2024). Because the rule-based layer operates deterministically, it introduces virtually no additional error when the input protocol is accurate. Essentially all final errors therefore originate at the Parser Agent's structuring stage, which is consistent with the Validator-dependent accuracy recovery observed in Fig. 7 and Fig. 8.

The hybrid design also aligns with the LLM-Modulo view that LLMs are unreliable as standalone planners but effective when combined with external verifiers (Kambhampati et al., 2024), and with reports that a deterministic constraint layer is needed for the safety of LLM-based robot agents (Yang et al., 2024). The rule-based engine follows this philosophy, deterministically enforcing physical constraints such as tip collision avoidance, plate position, and volume limits. The modular separation also benefits system integration. Although the output is instantiated as commands for a specific robotic laboratory platform, the intermediate structured protocol produced by the Parser and Validation Agents is system-agnostic. The same pipeline can therefore be ported to other automation platforms by replacing only the mapping rules of the rule-based engine, which provides a basis for future standardization of protocol compatibility across heterogeneous platforms.

### 4.5 Microplate-Based Liquid Handling and Total Protein Quantification Demonstration

Following the quantitative analyses above, this section examines how the proposed framework operates in actual microplate-based experiments through two demonstrations. In both cases, GPT-5 was used as the Parser Agent and Claude Sonnet 4.6 as the Validation Agent; the natural-language protocols passed validation on the 1st Pass without regeneration and were executed successfully on the robotic laboratory platform. The first demonstration is a liquid-handling case for a simple dispensing procedure. A single-sentence protocol — "Add 100 µL of Blank, Calibrant, or Sample to the appropriate wells. Incubate for 5 minutes at room temperature." — was provided as input, and the Parser Agent transformed it into a structured output of well positions, dispensing volumes, and incubation time. After passing the Validation Agent's checks for completeness, parameter consistency, and step ordering, the sequence was mapped by the rule-based engine to well-by-well dispensing commands, and the dispensing result on the plate matched the input protocol (Fig. 10(a)) The second demonstration is a Bradford assay measuring batch-to-batch variation in the total protein concentration of FBS. The input — "Pipette 10 µL of Blank (distilled water), Calibrant (5–25 µg/mL BSA), or serum sample into target wells. Add 200 µL of Bradford reagent per well. Shake at 300 rpm for 30 seconds. Incubate for 10 minutes at room temperature. Read absorbance at 595 nm and calculate total protein concentration against the standard curve." — is a five-step procedure: aspiration and dispensing, shaking, incubation, absorbance reading, and concentration calculation. The Parser Agent structured all quantitative parameters of each step — volumes, speeds, times, and wavelength — and the Validation Agent passed the output on the 1st Pass (Fig. 10(b)). Calibrated mean OD for the five-level BSA calibrant (5–25 µg/mL) increased monotonically from 0.5188 to 0.9948, forming the standard curve; the concentrations measured for the five diluted FBS samples ranged from 2.2143 to 4.4855 µg/mL, distinguishing the batches from one another (Fig. 10(c)). This shows that a naturally described ELISA-variant protocol can be converted automatically into analysis-ready results. That both demonstrations passed on the 1st Pass indicates that the GPT-5 and Claude Sonnet 4.6 combination produces reliable translation for standard microplate procedures with little use of the self-correction loop, consistent with the high 1st Pass rates of large Parser × large Validator combinations in Fig. 7 and Fig. 8.

These demonstrations illustrate how a researcher's workflow changes. Conventionally, researchers had to convert natural-language protocols into automation scripts, debug them, and execute them; with the present framework, the protocol serves as input, translation and verification are automatic, and the researcher handles only final confirmation and execution. This human-in-the-loop structure shifts the researcher's role from script writer to experimental designer and supervisor (Endsley, 2023), directing cognitive resources toward experimental design and hypothesis testing — the central aim of the self-driving laboratory. Several limitations warrant assessment. The experiments here were confined to microplate-based procedures centered on ELISA, dilution, and dispensing, and generalization to other experiment types — cell culture, PCR thermal cycling, centrifugation — requires separate validation. The modular design is intended to keep this cost at the level of incremental extension: new experiment types are added by extending the rule-based engine rather than redesigning the system. These limitations point to future directions. The most immediate is closed-loop automation: feeding experimental results, such as plate-reader measurements, back to the Validation Agent would let the system adjust subsequent conditions adaptively, extending the open-loop structure toward iterative experimental optimization (Boiko et al., 2023). For scientific validity, the verification criteria could incorporate domain ontology — assay-specific parameter ranges and reagent-compatibility rules — building on work that integrates external chemistry tools with LLMs (Bran et al., 2024). Further work includes extending the protocol scope to cell-culture and molecular-biology workflows and developing rule-based engines for additional automation platforms. Finally, refining the protocol dataset and evaluation framework from this study into a public benchmark for experimental automation would support reproducibility and comparability in follow-on research.

## 5. Conclusion

To bridge the semantic gap between unstructured natural-language microplate protocols and automation systems, this study proposed a dual-agent autonomous protocol translation framework combining an LLM-based Parser Agent, a rule-based mapping engine, and a heterogeneous LLM-based Validation Agent. The framework recovered the translation accuracy of mid- and small-sized Parsers to a level approaching that of large cloud models, and showed that the effect of cross-model verification depends not on generator–verifier heterogeneity alone but on the Validator's critical verification capability. Rule-based mapping outperformed direct LLM mapping in both accuracy and latency, supporting the hybrid architecture, and microplate liquid-handling and Bradford-assay demonstrations established end-to-end operation from natural-language protocol to real-world execution. The contributions are as follows: (i) integrating semantic structuring, deterministic command mapping, and heterogeneous-model cross-verification in a single framework that addresses the semantic gap of autonomous experimental systems; (ii) presenting a lightweight, model-agnostic path that raises the translation accuracy of mid- and small-sized LLMs toward that of large models using inference-time correction alone, without additional training or fine-tuning; and (iii) providing a model-selection guideline grounded in the accuracy–latency trade-off that supports adaptation to diverse deployment scenarios. Together, dual-agent correction, cross-model verification, and an LLM–rule hybrid architecture constitutes a practical, model-agnostic path for narrowing the gap between natural-language protocols and automation systems, bringing the vision of the self-driving laboratory closer by shifting researchers' cognitive resources from scripting toward experimental design and hypothesis testing.

## Acknowledgements

This work was supported by the Technology Innovation Program (2410012121, Development of Automated, Intelligent and Continuous Integration System for Biofoundry) and the Technology Innovation Program (2410016531, Integrated verification system for autonomous laboratory on linker – payload complexes) funded by the Ministry of Trade, Industry & Energy (MOTIE, Korea).

The authors used Claude Opus 4.7 (Anthropic, San Francisco, CA, USA) for language editing and translation during the preparation of this manuscript. The tool was not used to generate or interpret the study's data, results, or conclusions, and the authors reviewed and take full responsibility for the final text.

## Conflict of Interest

The authors declare no competing interests related to the research, authorship, or publication of this article.

**Figures and Tables**

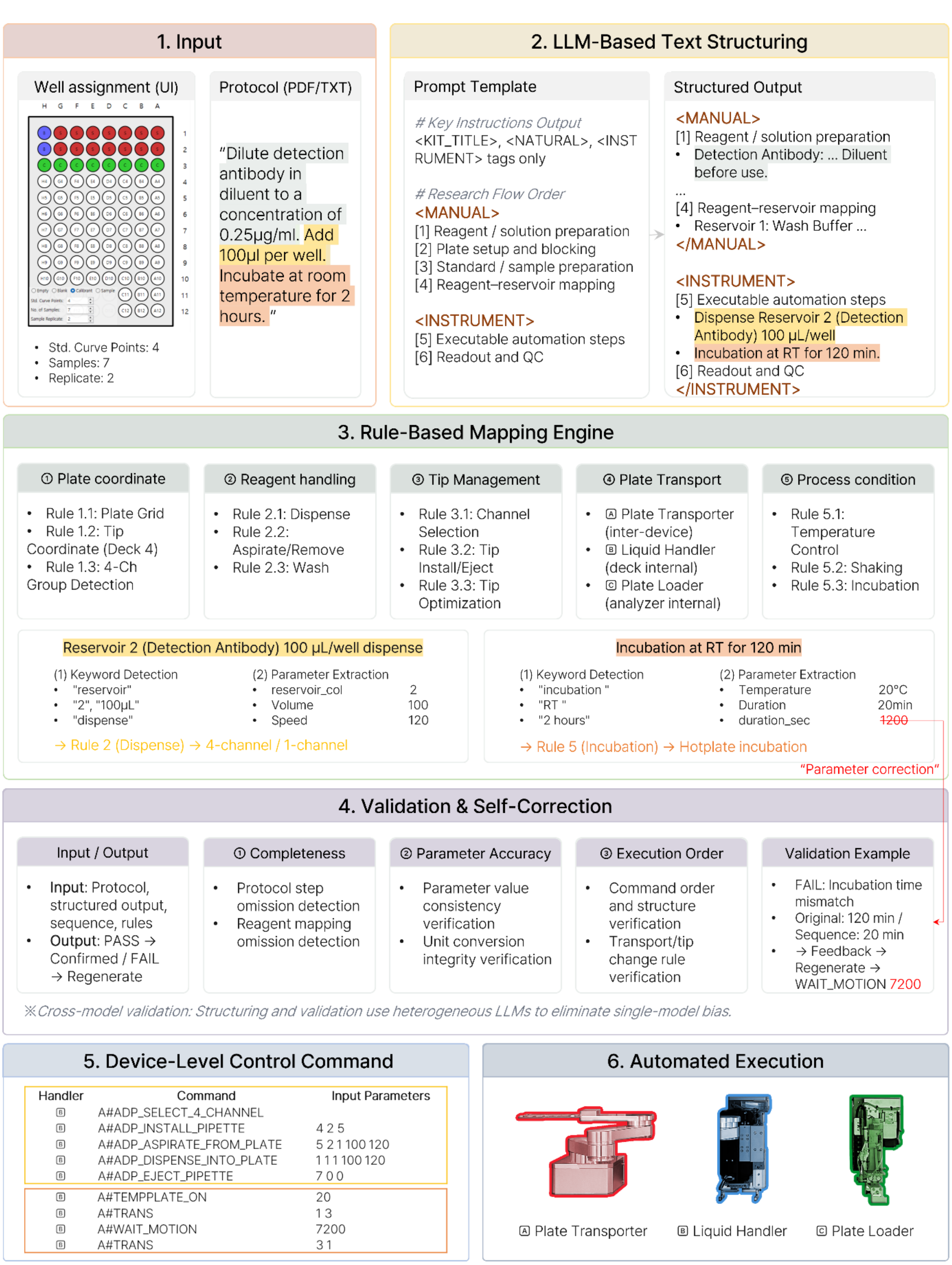

**Fig. 1. Overall pipeline of the proposed dual-agent protocol translation framework.** End-to-end pipeline diagram showing the transformation of a natural-language experimental protocol into a device-level control command sequence for the robotic laboratory platform, via semantic structuring by the Parser Agent, deterministic command mapping by the rule-based engine, and cross-model verification by the Validation Agent. Includes the feedback-based self-correction loop ($\leq 3$ iterations) triggered upon error detection by the Validation Agent, and the data flow by which only verified commands are transmitted to the automation platform.

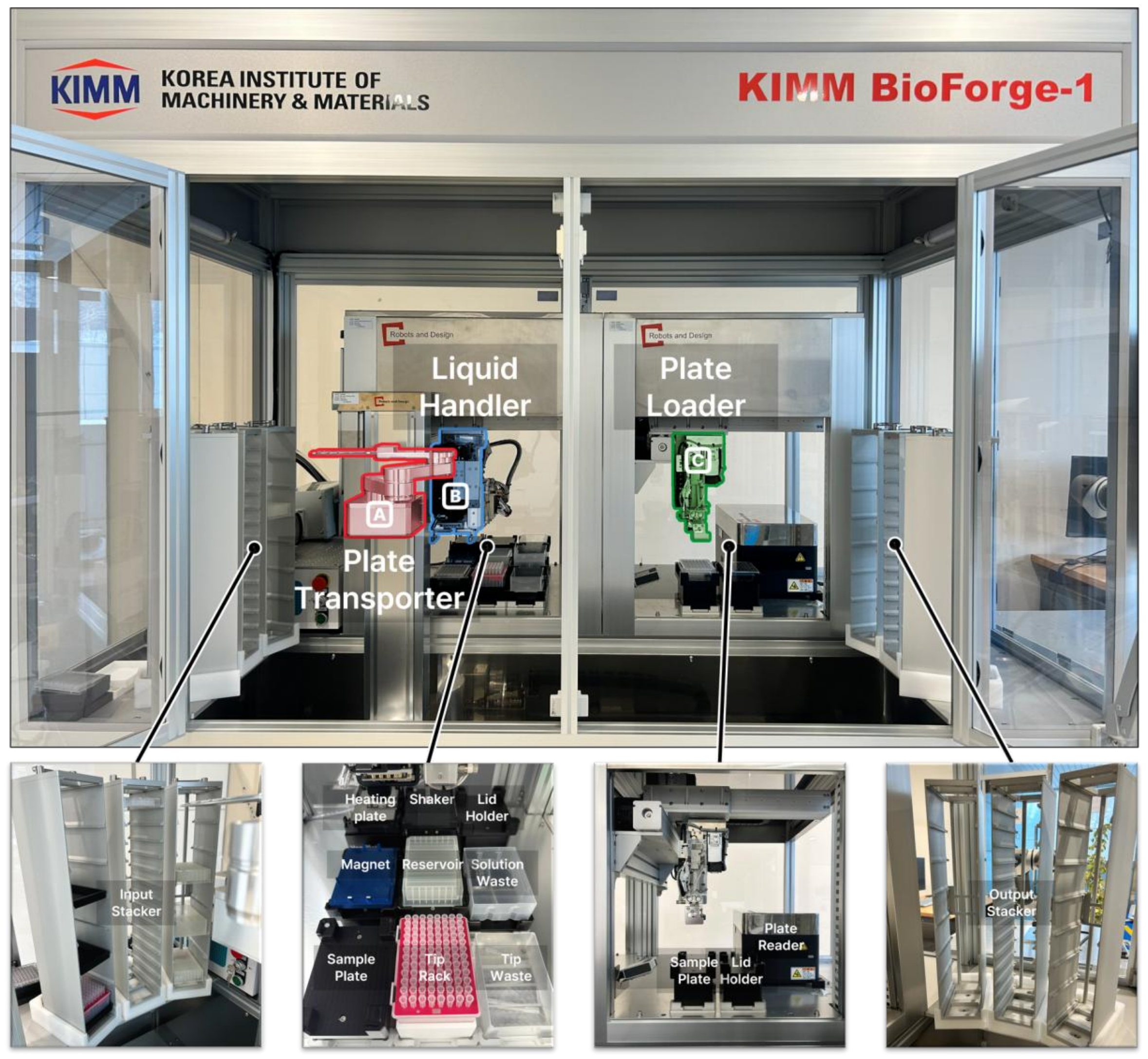

**Fig. 2. Robotic laboratory platform.** Hardware configuration of the robotic laboratory platform for automated execution of microplate-based biological experiments. Plate Transporter (A): microplate transport between the Main Deck and Analyzer Deck; Liquid Handler (B): liquid-handling operations including reagent aspiration, dispensing, tip installation, and washing; Plate Loader (C): intra-plate-reader movement and interface with analytical instruments. Unified integrated workflow environment comprising Input/Output Stacker, Sample Plate, Reagent Reservoir, Tip Rack, Tip Waste, Solution Waste, Shaker, Heating Plate, Magnet, Lid Holder, and Plate Reader modules.

**Input**

"PLATE PREPARATION Dilute capture antibody with PBS to a concentration of 0.125 µg/ml and immediately add 100 µl to each ELISA plate well, then seal the plate and incubate overnight at room temperature. Aspirate the wells to remove liquid and wash the plate 4 times using 300 µl of wash buffer per well...
ELISA PROTOCOL Standard/Sample: Dilute standard from 200 pg/ml to zero in diluent and immediately add 100 µl of standard or sample to each well in triplicate, incubate at room temperature for at least 2 hours. Detection: Aspirate and wash plate 4 times, dilute detection antibody to 0.25 µg/ml, add 100 µl per well... Streptavidin-HRP: Wash 4 times, add 100 µl per well, incubate 30 minutes... TMB substrate: Wash 4 times, add 100 µl substrate, incubate 20 minutes, add 100 µl 1M HCl stop solution. Read at 450 nm (620 nm correction)."

**Parser Agent Prompt**

**Instruction:**

[Data Extraction]
- Extract protocol steps and parameters from the input document.
- Preserve experimental intent and avoid unsupported additions.

[Output Format]
- <KIT_TITLE>, <MANUAL>,<INSTRUMENT>

[Workflow Order]
- <MANUAL>
  - [1] Reagent / solution preparation
  - [2] Plate setup and blocking
  - [3] Standard / sample preparation
  - [4] Reagent–reservoir mapping
- <INSTRUMENT>
  - [5] Executable automation steps
  - [6] Readout and QC

[Automation Rules]
- Separate user-performed and automated steps
- Extract all quantitative parameters (volume, time, temperature)
- Normalize values into execution-ready format
- Ensure compatibility with hardware constraints

[Reservoir Mapping]
- Assign fixed roles (e.g., wash buffer)
- Map reagents to reservoirs based on function and usage order

**Structured Output**

<MANUAL>
[1] Reagent / solution preparation (user)
- Dilute capture antibody to 0.125 µg/mL,...

[2] Plate setup and blocking (user)
- Wash 4× with 300 µL/well,...

[3] Standard / sample preparation(user)
- Dilute standard (200 → 0 pg/mL),...

[4] Reagent–reservoir mapping
- Reservoir 1: Wash Buffer
- Reservoir 2: Detection Antibody (0.25 µg/mL in diluent)
- Reservoir 3: Streptavidin-HRP (0.10 µg/mL in diluent)
- Reservoir 4: TMB Substrate
- Reservoir 5: 1M HCl Stop Solution

</MANUAL>

<INSTRUMENT>
[5] Executable automation steps
- Aspirate Standard/Sample 100 µL/well
- Wash 4× with Reservoir 1, 300 µL/well
- Dispense Reservoir 2, 100 µL/well
- Incubate 2 hours at RT

...
- Wash 4× with Reservoir 1, 300 µL/well
- Dispense Reservoir 4, 100 µL/well
- Incubate 20 minutes at RT
- Dispense Reservoir 5, 100 µL/well

[6] Readout and QC
- Read at 450 nm (620 nm correction)

</INSTRUMENT>

**Fig. 3. Parser Agent for LLM-based structuring of natural-language protocols.** The Parser Agent applies a constrained interpretation prompt to convert an unstructured natural-language protocol (left) into a tagged structured representation (right). Steps are separated into two domains by the performing agent: <MANUAL> for user-executed steps and <INSTRUMENT> for instrument-executed steps. The example shown is an ELISA protocol.

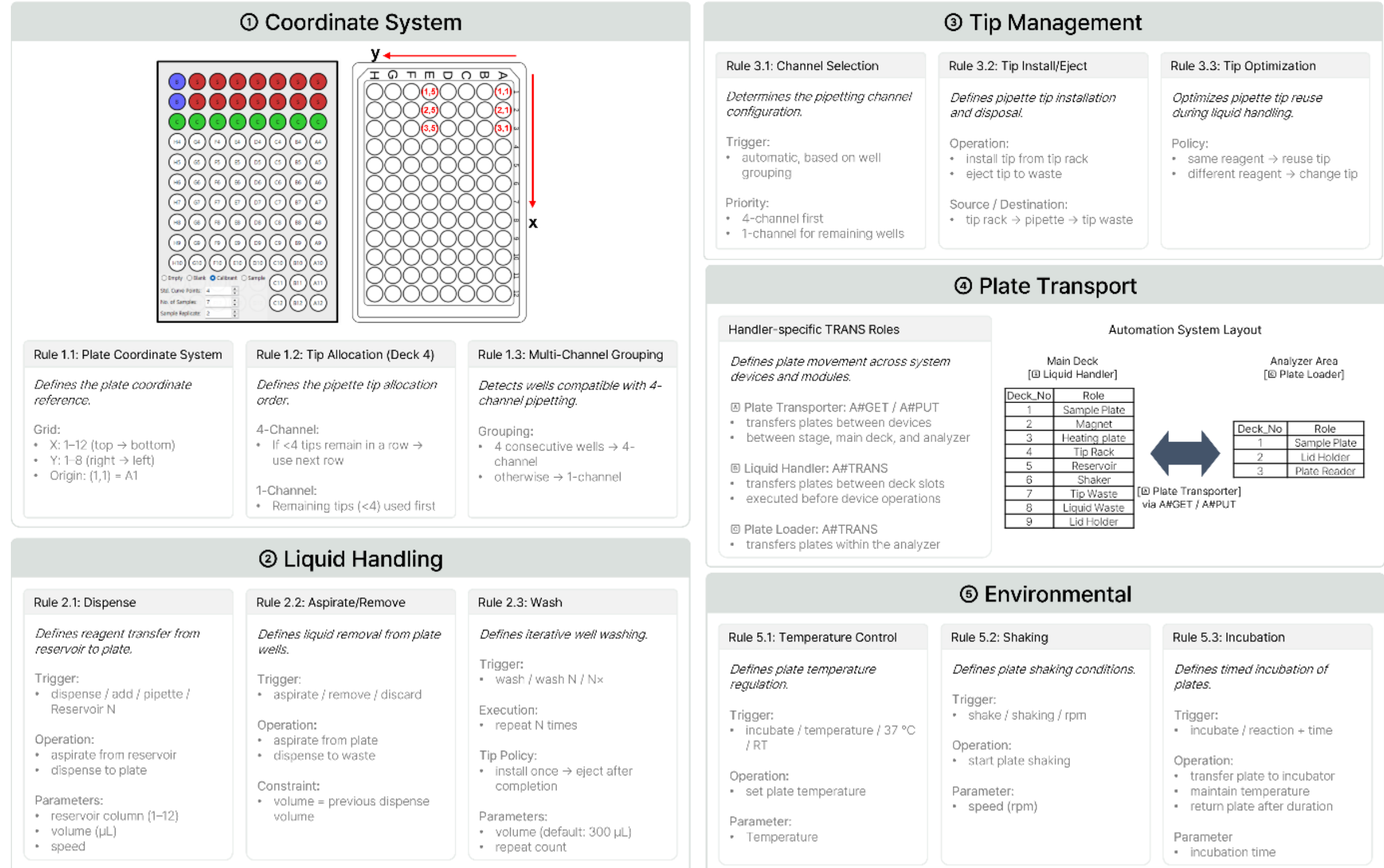

**Fig. 4. Architecture and rule layers of the rule-based command mapping engine.** The engine operates as a deterministic execution layer that maps a structured protocol to device-level control commands through a five-tier rule hierarchy: (1) Coordinate System, (2) Liquid Handling, (3) Tip Management, (4) Plate Transport, and (5) Environmental Control. Each tier defines sub-rules governing the corresponding aspect of instrument operation. Handler-specific transport roles for the Plate Transporter, Liquid Handler, and Plate Loader are also shown.

**1. Structured Protocol**

| Step | Structured Protocol |
|---|---|
| [1] | Remove previous solution from all wells. |
| [2] | Wash 4 times with 300 μL Wash Buffer. |
| [3] | Dispense 100 μL Detection Antibody from Reservoir 2. |
| [4] | Incubate at room temperature for 2 hours. |
| [5] | Read absorbance at 450 nm. |

**2. Rule Assignment**

| Step | Trigger | Applied Rules | Extracted Parameters |
|---|---|---|---|
| [1] | remove, solution | 2.2, 3.1, 1.3 | vol=100μL, dest=Slot 8 |
| [2] | wash, 4 times, 300μL | 2.3 | repeat=4, vol=300μL, reservoir=1 |
| [3] | dispense, 100μL, Reservoir 2 | 2.1, 3.3 | vol=100μL, reservoir=2 |
| [4] | incubate, room temperature, 2 hours | 5.1, 5.3 | temp=20 °C, time=7200s |
| [5] | read, absorbance, 450nm | 4-(A)(B)(C) | path: B→A→C |

**3. Device-Level Control Command Output**

[1] Remove (14 commands)
SELECT_4_CHANNEL → INSTALL → ASPIRATE(plate) → DISPENSE(waste) ×2 → EJECT
SELECT_1_CHANNEL → INSTALL → ASPIRATE(plate) → DISPENSE(waste) ×3 → EJECT

[2] Wash 4× (34 commands)
SELECT_4_CHANNEL → INSTALL → [ASPIRATE(Reservoir1)→ DIS(plate)→ ASP(plate)→ DIS(waste)] ×4 ×2groups → EJECT
SELECT_1_CHANNEL → INSTALL → [ASPIRATE(Reservoir1)→ DISPENSE(plate)→ ASP(plate)→ DIS(waste)] ×4 ×3wells → EJECT

[3] Dispense (14 commands)
SELECT_4_CHANNEL → INSTALL → [ASPIRATE(Reservoir2)→ DISPENSE(plate)] ×2groups → EJECT
SELECT_1_CHANNEL → INSTALL → [ASPIRATE(Reservoir2)→ DISPENSE(plate)] ×3wells → EJECT

[4] Incubate (4 commands)
TRANS(1→3) → TEMPPLATE_ON(20) → WAIT(7200) → TRANS(3→1)

[5] Read OD (12 commands)
(B) CAP_CLOSE
(A) GET(4,1) → PUT(5,1)
(C) CAP_OPEN → ANALYZER_OPEN → TRANS(1→3) → START → TRANS(3→1) → CLOSE
(A) GET(5,1) → PUT(2,1)

**Fig. 5. Expansion of structured-protocol steps into device-level command sequences.** Each step of the structured protocol (Top left) is mapped to a rule assignment (Top right) and then expanded into a sequence of device-level control commands (bottom). The example shows that a single semantic step can decompose into several to several dozen commands, highlighting the granularity gap that the rule-based engine absorbs between semantic units and execution units.

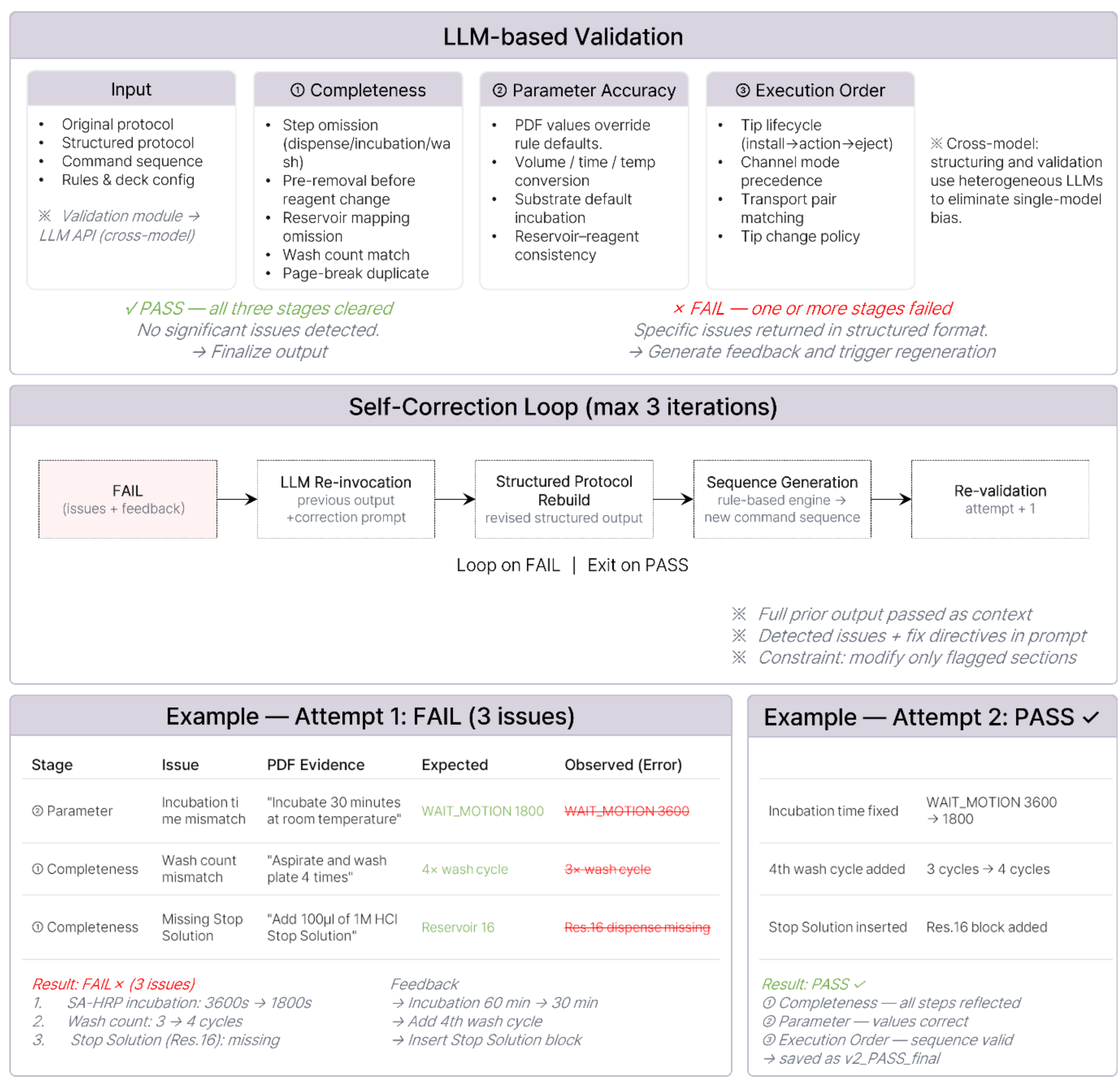

**Fig. 6. Validation Agent and self-correction loop with worked example.** Top: the Validation Agent receives the original protocol, structured protocol, and command sequence, and verifies them against three criteria — Completeness, Parameter Accuracy, and Execution Order — using a heterogeneous LLM distinct from the generation model to avoid single-model bias. Middle: when verification fails, structured feedback drives a self-correction loop (up to three iterations) that regenerates the structured protocol and command sequence before re-validation. Bottom: a worked ELISA example in which Attempt 1 fails on three issues (incubation duration, wash count, missing Stop Solution) and Attempt 2 passes after correction.

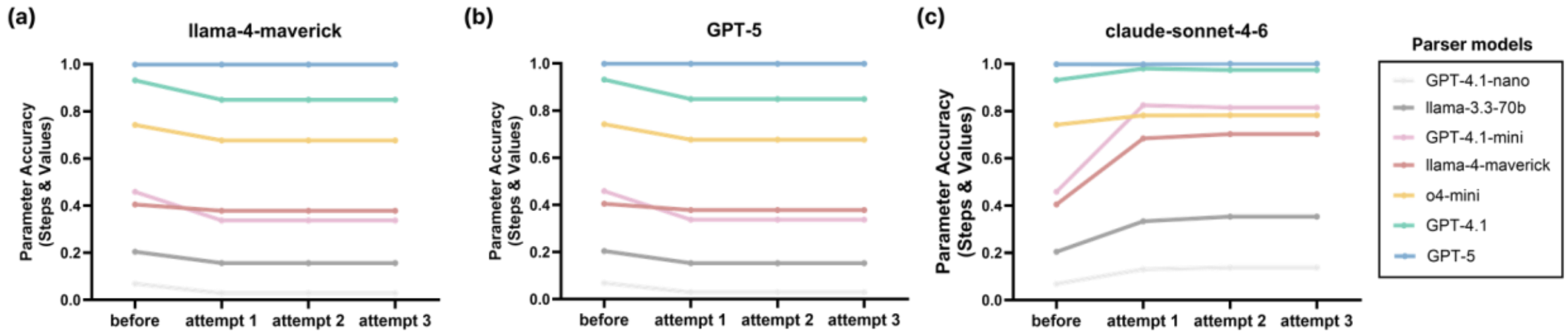

**Fig. 7. Effectiveness of iterative correction across Validators.** Graphs of parameter accuracy across w/o regen, attempt 1, attempt 2, and attempt 3 stages for seven Parser models (GPT-5, GPT-4.1, o4-mini, llama-4-maverick, GPT-4.1-mini, llama-3.3-70b, GPT-4.1-nano) under each of three Validation Agents. (a) Validator: llama-4-maverick; (b) GPT-5; (c) claude-sonnet-4-6.

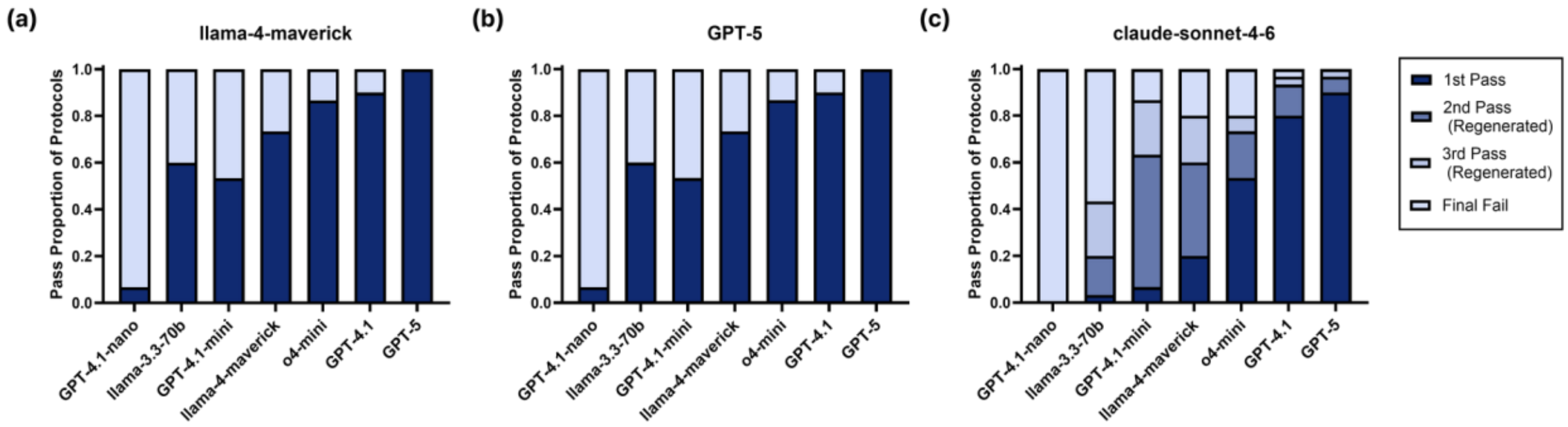

**Fig. 8. Final pass rate decomposition under cross-verification.** Pass proportion of protocols for each Parser × Validator combination across 30 protocols, decomposed into four categories: 1st Pass, 2nd Pass (Regenerated), 3rd Pass (Regenerated), and Final Fail. (a) Validator: llama-4-maverick; (b) GPT-5; (c) claude-sonnet-4-6.

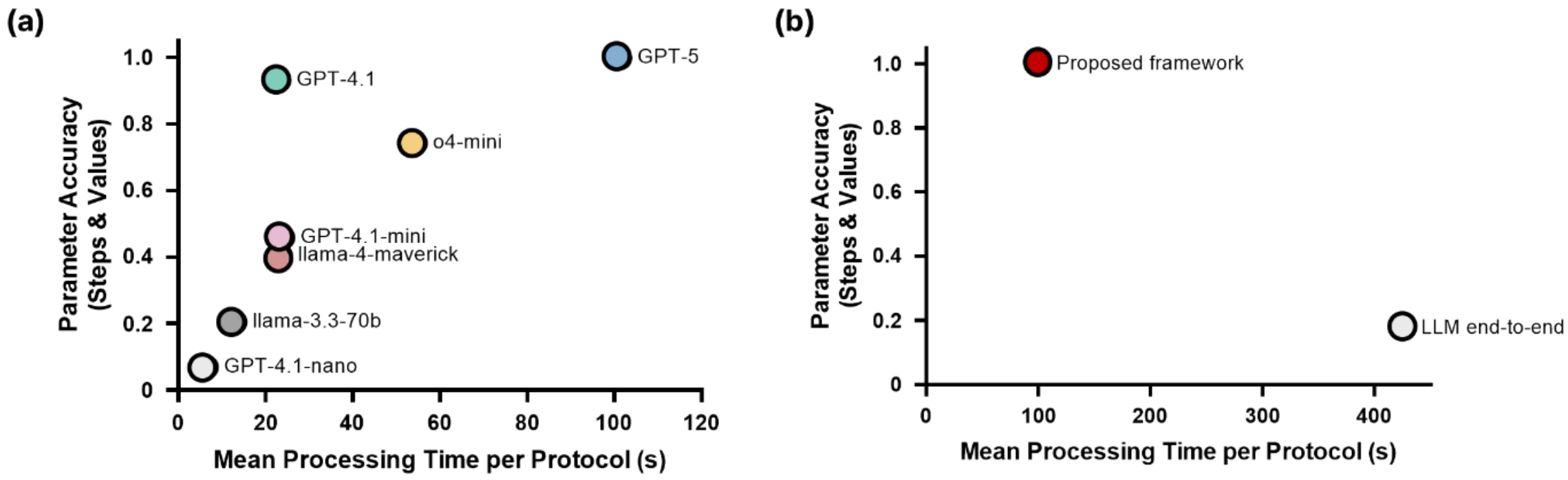

**Fig. 9. Accuracy–latency comparison of Parser Agents and of the proposed framework against LLM end-to-end mapping.** Parameter accuracy is plotted against mean processing time per protocol. (a) Position of each Parser Agent in the accuracy–latency plane without the self-correction loop (w/o regen.), showing the baseline trade-off between accuracy and processing time. (b) Comparison between the proposed framework (LLM-based structuring + rule-based command mapping) and an LLM end-to-end baseline that directly outputs device-level control commands; the same GPT-5 model is used in both conditions to isolate the effect of the rule-based mapping engine.

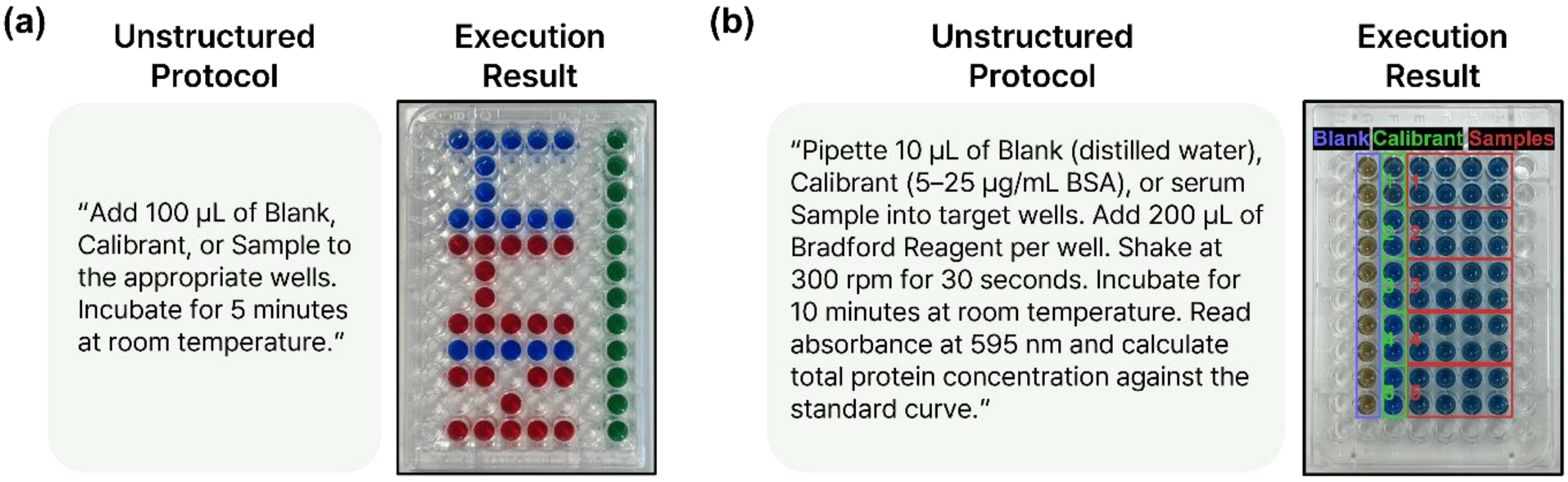

※ Parser Agent: GPT-5 | Validation Agent: Claude Sonnet 4.6 | 1st Pass

(c)

| | OD Average (corrected) | Concentration (µg/ml) |
|---|---|---|
| Calibrant 1 | 0.5188 | 5 (Known) |
| Calibrant 2 | 0.7383 | 10 |
| Calibrant 3 | 0.8408 | 15 |
| Calibrant 4 | 0.9073 | 20 |
| Calibrant 5 | 0.9948 | 25 |

| | OD Average (corrected) | Concentration (µg/ml) |
|---|---|---|
| Sample 1 | 0.8154 | 4.4855 (Calculated) |
| Sample 2 | 0.7880 | 3.2634 |
| Sample 3 | 0.7771 | 2.7779 |
| Sample 4 | 0.7645 | 2.2143 |
| Sample 5 | 0.7699 | 2.4542 |

**Fig. 10. End-to-end execution of microplate experiments using the proposed framework.** Demonstrations performed on the robotic laboratory platform with GPT-5 as the Parser Agent and Claude Sonnet 4.6 as the Validation Agent, both completed on the first attempt. (a) Liquid-handling demonstration: a single-sentence protocol is structured and executed to dispense Blank, Calibrant, and Sample into designated wells. (b) Bradford assay for total protein quantification in fetal bovine serum (FBS), executed end-to-end from protocol parsing through dispensing, shaking, incubation, and absorbance measurement. (c) Standard curve from the five-point BSA calibrant (5–25 µg/mL) and the total protein concentrations measured for the five diluted FBS samples.

**Table 1. Large language models used as Parser and Validator agents.** Seven models were evaluated as Parser Agents (GPT-5, GPT-4.1, o4-mini, GPT-4.1-mini, GPT-4.1-nano, Llama-4-Maverick, Llama-3.3-70B) and three as Validator Agents (Claude Sonnet 4.6 (default), GPT-5, Llama-4-Maverick). All model versions and release dates are as of 2026-05-22.

| **Model** | **Role in study** | **Provider** | **Hosting** | **Version / snapshot** | **Release date** |
|---|---|---|---|---|---|
| GPT-5 | Parser; Validator | OpenAI | Cloud API | gpt-5 (gpt-5-2025-08-07) | 2025-08-07 |
| GPT-4.1 | Parser | OpenAI | Cloud API | gpt-4.1-2025-04-14 | 2025-04-14 |
| o4-mini | Parser | OpenAI | Cloud API | o4-mini-2025-04-16 | 2025-04-16 |
| GPT-4.1-mini | Parser | OpenAI | Cloud API | gpt-4.1-mini-2025-04-14 | 2025-04-14 |
| GPT-4.1-nano | Parser | OpenAI | Cloud API | gpt-4.1-nano-2025-04-14 | 2025-04-14 |
| llama-4-maverick | Parser; Validator | Meta | Local / open-weight | Llama-4-Maverick-17B-128E-Instruct | 2025-04-05 |
| llama-3.3-70b | Parser | Meta | Local / open-weight | Llama-3.3-70B-Instruct | 2024-12-06 |
| claude-sonnet-4-6 | Validator (default) | Anthropic | Cloud API | claude-sonnet-4-6 | 2026-02-17 |

## Supplementary Information

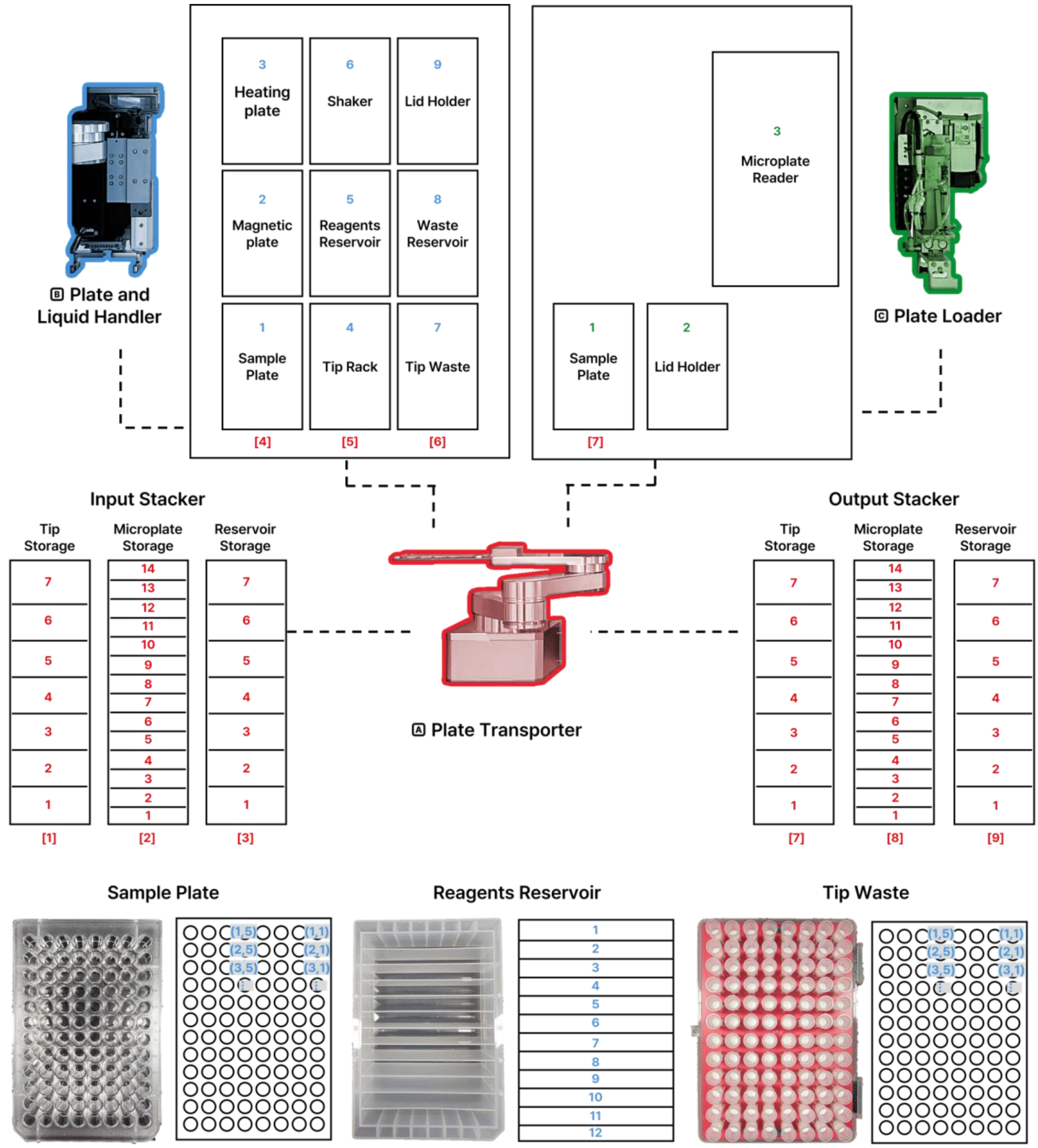

**Supplementary Fig. 1. Coordinate and labeling system of the robotic laboratory platform.** (A) Plate Transporter area, showing tip storage, microplate storage, and reservoir storage in the Input and Output Stackers. (B) Liquid Handler area, comprising nine deck slots for sample plate, tip rack, tip waste, magnetic plate, reagents reservoir, waste reservoir, heating plate, shaker, and lid holder. (C) Plate Loader area, including sample plate, lid holder, and microplate reader positions. The bottom panels show the coordinate notation (e.g., (1,1)–(3,5)) for the Sample Plate, Reagents Reservoir, and Tip Waste, which defines the mapping between physical well locations and equipment coordinates used by the rule-based mapping engine.